\pdfoutput=1

\documentclass[11pt]{article}

\usepackage[]{acl}

\usepackage{times}
\usepackage{graphicx}
\usepackage{latexsym}
\usepackage{amsfonts}
\usepackage{amsmath}
\usepackage{MnSymbol}
\usepackage{graphicx}
\usepackage{booktabs}
\usepackage{bbding}
\usepackage{makecell}
\usepackage{multirow}
\usepackage{tabularx}
\usepackage{color}
\definecolor{green}{RGB}{68,169,32}
\definecolor{olive}{RGB}{128,128,0}
\definecolor{purple}{RGB}{102,51,102}
\definecolor{pink}{RGB}{255,20,147}
\definecolor{blue}{RGB}{0,0,139}

\usepackage{float}

\usepackage[T1]{fontenc}

\usepackage[utf8]{inputenc}

\usepackage{microtype}

\title{UniDU: Towards A Unified Generative Dialogue Understanding Framework}

\author{Zhi Chen$^{1}$, Lu Chen$^{1}$\thanks{The corresponding authors are Lu Chen and Kai Yu.}, Bei Chen$^{2}$, Libo Qin$^{2}$, Yuncong Liu$^{1}$,
\\ \textbf{Su Zhu}$^{3}$, \textbf{Jian-Guang Lou}$^{2}$ \and \textbf{Kai Yu}$^{1}$\footnotemark[1] \\
        $^1$X-LANCE Lab, Department of Computer Science and Engineering \\ 
        MoE Key Lab of Artificial Intelligence, AI Institute, Shanghai Jiao Tong University\\
    State Key Lab of Media Convergence Production Technology and Systems, Beijing, China\\
    $^2$Microsoft Research Asia \\
    $^3$AISpeech Co., Ltd., Suzhou, China \\
        }

\begin{document}
\maketitle
\begin{abstract}

With the development of pre-trained language models, remarkable success has been witnessed in dialogue understanding (DU). However, current DU approaches usually employ independent models for each distinct DU task without considering shared knowledge across different DU tasks. In this paper, we propose a unified generative dialogue understanding framework, named {\em UniDU}, to achieve effective information exchange across diverse DU tasks. Here, we reformulate all DU tasks into a unified prompt-based generative model paradigm. More importantly, a novel model-agnostic multi-task training strategy (MATS) is introduced to dynamically adapt the weights of diverse tasks for best knowledge sharing during training, based on the nature and available data of each task. Experiments on ten DU datasets covering five fundamental DU tasks show that the proposed UniDU framework largely outperforms task-specific well-designed methods on all tasks. MATS also reveals the knowledge-sharing structure of these tasks. Finally, UniDU obtains promising performance in the unseen dialogue domain, showing the great potential for generalization.

\end{abstract}

\section{Introduction}
\label{sec:intro}
The development of the conversational system plays an important role in the spread of the intelligence devices, such as intelligence assistants and car play. In recent years, there has been a growing interest in neural dialogue system~\cite{wen2017network,ultes2017pydial,li2017end,chen2018policy,chen2019agentgraph,chen2020distributed,bao2020plato,adiwardana2020towards,ham2020end,peng2020soloist,chen2022dialogzoo}. 
Dialogue understanding is a core technology and hot topic in the dialogue system, aiming to analyze a dialogue from different fine-grained angles accurately.

There are five classical dialogue understanding tasks: dialogue summary (DS)~\cite{liu2019automatic}, dialogue completion (DC)~\cite{su2019improving,quan2020risawoz}, intent detection (ID)~\cite{kim2016intent,casanueva2020efficient,qin2021co}, slot filling (SF)~\cite{zhang2017position,qin2021survey,haihong2019novel} and dialogue state tracking (DST)~\cite{kim2020efficient,chen2020schema,hosseini2020simple,xu2020memory,liao2021dialogue}. 
Dialogue summary aims to generate a concise description of given dialogue content, which is normally formulated as a sequence-to-sequence generation problem~\cite{wu2021controllable}. Dialogue completion eliminates the co-reference and information ellipsis in the latest utterance, which is also a generation task~\cite{Chen2021DecoupledDM}. Intent detection and slot filling are two traditional spoken language understanding tasks that aim to map natural language to logical form. Intent detection is typically treated as a classification problem~\cite{liu2016attention} and slot filling is usually formulated as a sequence labeling task~\cite{zhang2017position,qin2019stack,coope2020span}. The dialogue state tracking task is to extract the user's constraints on the predefined dialogue domains and slots~\cite{budzianowski2018multiwoz}.
The five different tasks aim to interpret a dialogue from five different perspectives. To date, these DU tasks are still learned independently due to different task formats. However, they are intuitively related. For example, the dialogue completion task should have a positive effect on the dialogue state tracking task~\cite{han2020multiwoz}. On the other hand, it is usually very expensive to collect dialogue data and annotate them, which constraints the scale of annotated dialogue corpora. It is important and imperative to study how to enhance dialogue understanding capability with the existing diverse dialogue corpora.


There are two main challenges in knowledge sharing across DU tasks: {\em data annotation} diversity and {\em task nature} diversity. It is necessary to employ a unified DU model to allow all types of DU data to be used together. In this paper, we propose a  \textbf{Uni}fined \textbf{D}ialogue \textbf{U}derstanding (UniDU) framework, in which the five fundamental DU tasks are modelled by a unified sequence-to-sequence generative model. The second challenge is related to the nature of diverse tasks. Since the output label dynamic ranges and the goals of the DU tasks are different, tasks may not be well suited to be trained together with straightforward multi-task learning. It is then a nontrivial problem to effectively weight diverse tasks for the unified model with different dialogue corpora. In this paper, we propose a novel adaptive weighting approach and compare it with other different training strategies under the UniDU framework.

The main contributions of this paper are summarized below:
\begin{itemize}
    \item To the best of our knowledge, we are the first to formulate different dialogue understanding tasks as a unified generation task spanned five DU tasks. The proposed UniDU outperforms well-designed models on five well-studied dialogue understanding benchmarks.
    \item We propose a model-agnostic adaptive weighting approach for multitask learning to address the task nature diversity problem. We find that the intuitive multitask mixture training method makes the unified model bias convergence to more complex tasks. The proposed model-agnostic training method can efficiently relieve this problem. 
    \item Experimental results show that the proposed UniDU method has excellent generalization ability, which achieves advanced performance both on few-shot and zero-shot setups.
\end{itemize}

\section{Dialogue Understanding Tasks}
We denote dialogue context as $C=(H_n,U_n)$, where $H_n=(U_1,U_2,\dots,U_{n-1})$ represents the dialogue history containing the first $n-1$ turns of utterances. $U_n$ is $n$-th turn utterance, which may consist of multiple sentences stated by one speaker. For the task-oriented dialogue, the domain scope is restricted by the dialogue ontology, which the dialogue expert designs. The ontology $O$ is composed of dialogue domains $D=\{d\}$ (like \emph{hotel}), domain slots (like \emph{price}) $S=\{s\}$ and user intent candidates $I=\{i\}$ (like \emph{find\_hotel}). There are five fundamental tasks to interpret a dialogue from different perspectives.

\noindent \textbf{Dialogue Summary} (DS) aims to extract important information of the dialogue. It is a typical generation problem, which takes the whole dialogue context $C$ as input and generates the summary description. DS requires the model to focus on the whole dialogue flow and the important concepts.

\noindent \textbf{Dialogue Completion} (DC) purposes to relieve the co-reference and information ellipsis problems, which frequently occur in the dialogue context. It is also a typical generation task, which inputs the dialogue history $H_n$ and the current utterance $U_n$ and then infers the semantic-completed statement of the current utterance $U_n$. DC requires the model to focus on the connection between current utterance and dialogue history.

\noindent \textbf{Slot Filling} (SF) is to extract the slot types $S$ of the entities mentioned by the user. It is a word tagging problem where the utterance is labeled in the IOB (Inside, Outside, and Beginning) format. The input is only the current utterance $U_n$.

\noindent \textbf{Intent Detection} (ID) is to recognize the intent from predefined abstracted intent expresses $I$. It is normally formulated as a classification problem. The input is the current utterance $U_n$, and the output is the possible distribution of all the intent candidates $I$.

\noindent \textbf{Dialogue State Tracking} (DST) aims to record the user's constraints, which consists of the triple set of domain-slot-value. For example, hotel-price-cheap means the user wants a cheap hotel. The input of DST at the $n$-th turn is the first $n$ turns $(U_1,\dots,U_n)$.

\begin{figure*}[t]
\centering
\includegraphics[width=0.9\textwidth]{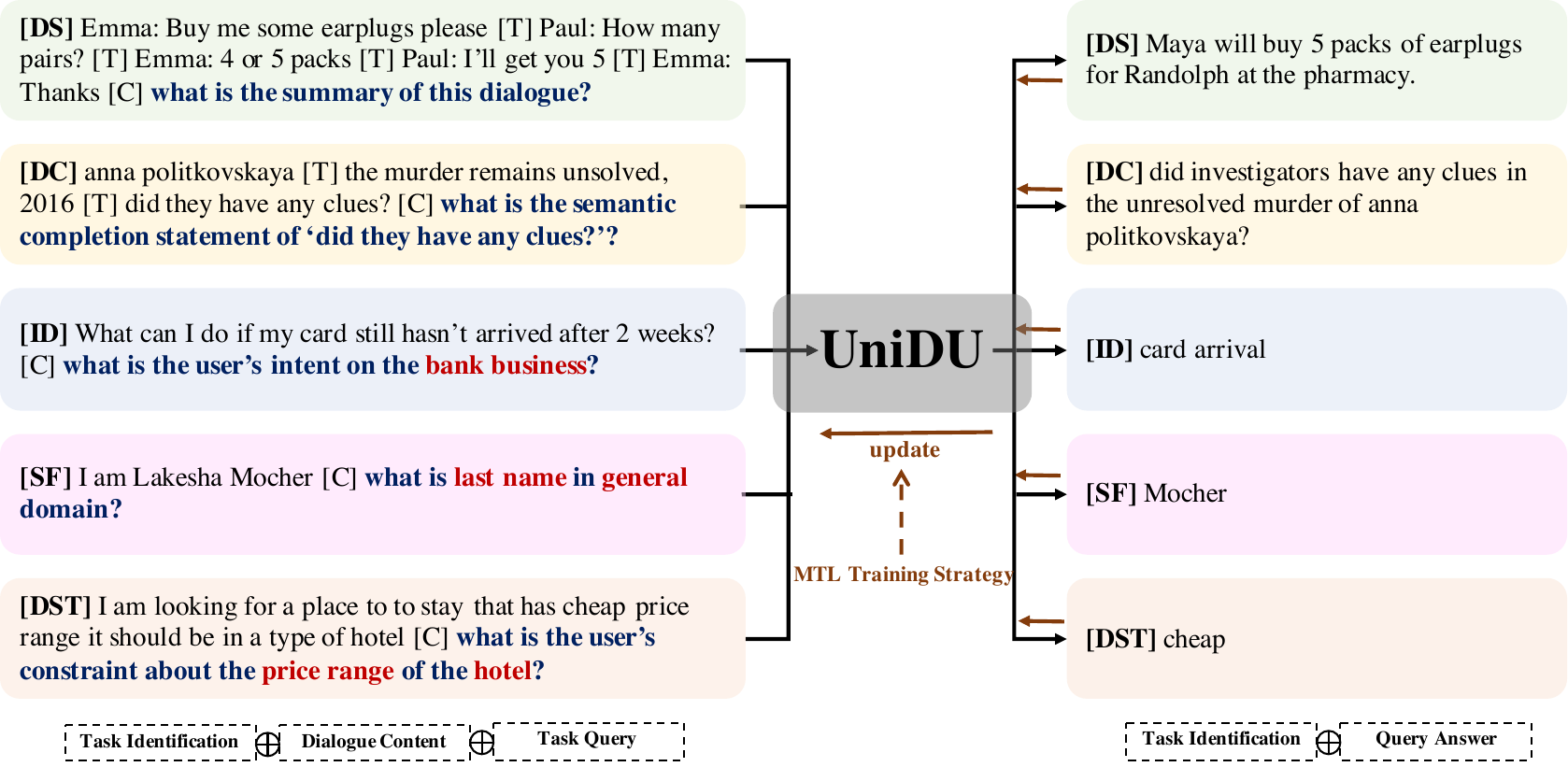} 
\caption{Overview of UniDU. Under UniDU framework, the input consists of three parts: task identification, dialogue content and task query, where $\oplus$ means concatenation. The output has two components: task identification and query answer. We train the UniDU model with different multitask learning strategies.}
\label{unidu}
\vspace{-3mm}
\end{figure*}

\section{UniDU}
\label{sec:unidu}
In this section, we first introduce the unified sequence-to-sequence data format for the five DU tasks. Then we introduce the formulation of each task in detail, especially how to reformulate the intent detection, slot filling and dialogue state tracking as the generation task.

There are three components in the input of UniDU: task identification, dialogue content, and task query. The task identification represents with a special token, e.g., dialogue summary identified by ``[DS]''. The dialogue content means the task-dependent input, such as dialogue history for dialogue summary. The task query can be regarded as the task-specific prompt, which includes the task definition and domain-related information. There are two elements in the output of UniDU: task identification and query answer. The query answer is the understanding result of the task query given by the dialogue content. The unified input and output can be formalized as:
\begin{align*}
\begin{tabular}{ll}
    \textbf{I{\small NPUT}}: & [TI] dialogue content [C] task query \\
    \textbf{O{\small UTPUT}}: & [TI] query answer
\end{tabular}
\end{align*}
\noindent where ``[C]'' is separate character and ``[TI]'' is task identification (replaced by ``[DS]'', ``[DC]'', ``[SF]'',``[ID]'' and ``[DST]'', which correspond to dialogue summary, dialogue completion, slot filling, intent detection and dialogue state tracking respectively). At inference time, the UniDU model must first predict the task identification.

Dialogue summary and dialogue completion are originally generative tasks. The dialogue contents in the input are the whole dialogue context $C$ and multi-turn utterances $H_n$ respectively. Since these two tasks are independent of the dialogue domain, there is no domain information in the task query. For dialogue summary, the task query is ``\emph{what is the summary of this dialogue?}''. For dialogue completion, the query is `\emph{`what is the semantic completion statement of $U_n$?}'', where $U_n$ is the $t$-th utterance. Their understanding answers are annotated dialogue summaries and rewritten utterances in the output.

The original slot filling task demands the model to extract all the mentioned slot values and their slot types in an utterance $U_n$. In this paper, the UniDU model predicts the value slot by slot, which is an iterated generation process on the slot candidate list. Two different slot filling formats are shown below:
\begin{figure}[H]
\centering
\includegraphics[width=0.45\textwidth]{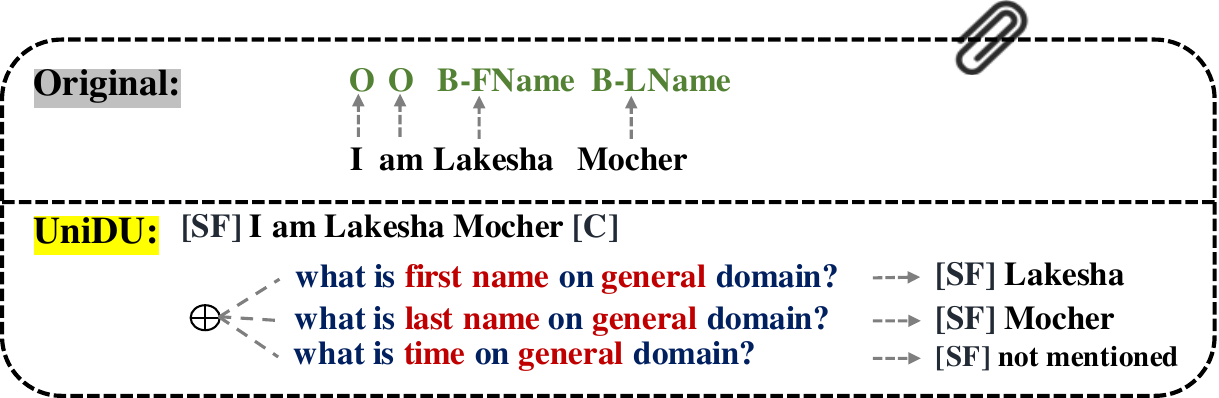} 
\end{figure}
\noindent To be clear, we do not list all the candidate slots here. 
In general, for each sample, it can be formalized as:
\begin{align*}
\begin{tabular}{ll}
    \textbf{I{\small NPUT}}: & [SF] $U_n$ [C] what is $s$ of $d$?  \\
    \textbf{O{\small UTPUT}}: & [SF] slot value
\end{tabular}
\end{align*}
where $s$ and $d$ are predefined slots and domains. If $s$ has no value in $U_n$, slot value will be ``not mentioned''. If $s$ has multiple values, they will be separated by a comma in the slot value. When the value is ``not mentioned'', we call it a negative sample. Otherwise, it is a positive sample. To balance the ratio of negative and positive samples during the training process, we set the ratio to 2:1. If the number of negative samples exceeds the threshold, we randomly sample twice as many negative instances as positive ones. 

For dialogue state tracking tasks, the classification methods always achieve better performance than generative methods. However, under the UniDU framework, we also formulate DST as a slot-wise value generation task similar to the slot filling task. The DST task formats are shown below:
\begin{figure}[H]
\centering
\includegraphics[width=0.45\textwidth]{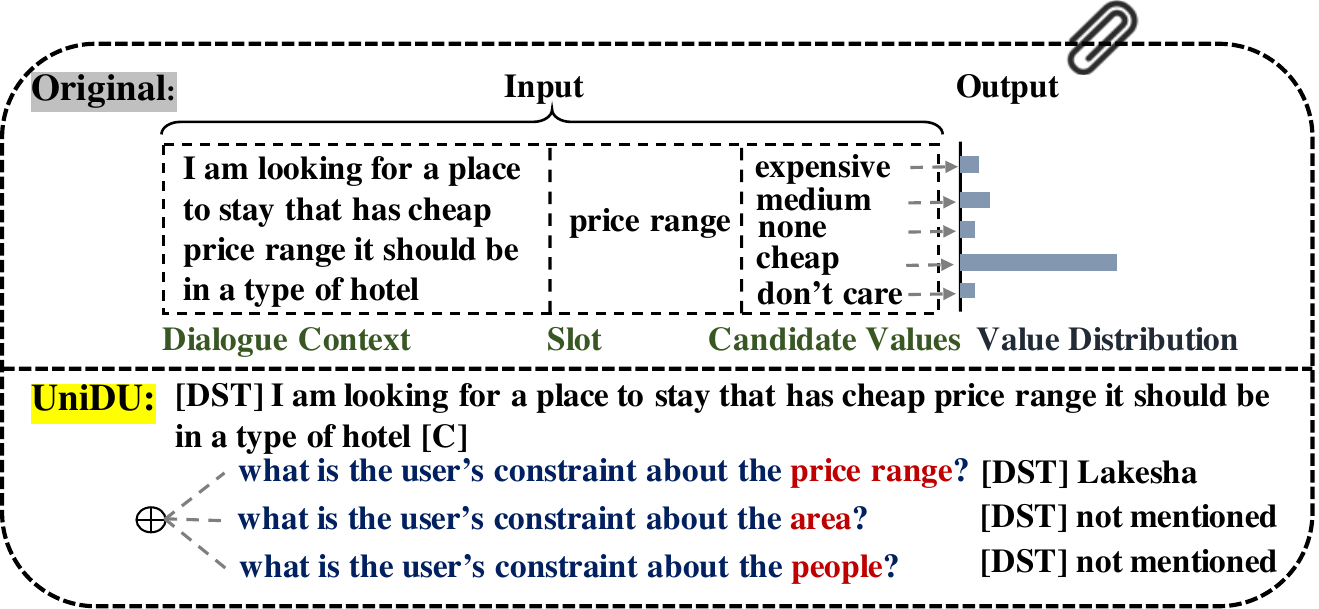} 
\end{figure}

\noindent where the output of the original DST model is the distribution of all the candidate values of the slot.
The input and output of the DST task under UniDU can be formalized as follows:
\begin{align*}
\begin{tabular}{ll}
    \textbf{I{\small NPUT}}: & [DST] $(H_n,U_n)$ [C] what is the \\
    & user's constraint about $s$ of $d$?  \\
    \textbf{O{\small UTPUT}}: & [DST] slot value
\end{tabular}
\end{align*}
where $(H_n,U_n)$ is dialogue context. If slot $s$ of the domain $d$ is not in the dialogue state, its value is ``not mentioned'', which is a negative sample. Note that different utterances are separated by the special token ``[T]'' in the input. During the training process, the ratio of negative and positive samples is also set below 2:1.

For the intent detection task, the original methods formulate it as the intent classification problem and output the distribution of all the candidate intents.
The UniDU model directly generates the intent name of the current utterance, which can be formalized as:
\begin{align*}
\begin{tabular}{ll}
    \textbf{I{\small NPUT}}: & [ID] $U_n$ [C] what is the user's intent \\
    & on domain $d$?  \\
    \textbf{O{\small UTPUT}}: & [ID] intent name
\end{tabular}
\end{align*}
where domain $d$ is normally known in advance. The specific examples of original and UniDU formats are shown below:
\begin{figure}[H]
\centering
\includegraphics[width=0.45\textwidth]{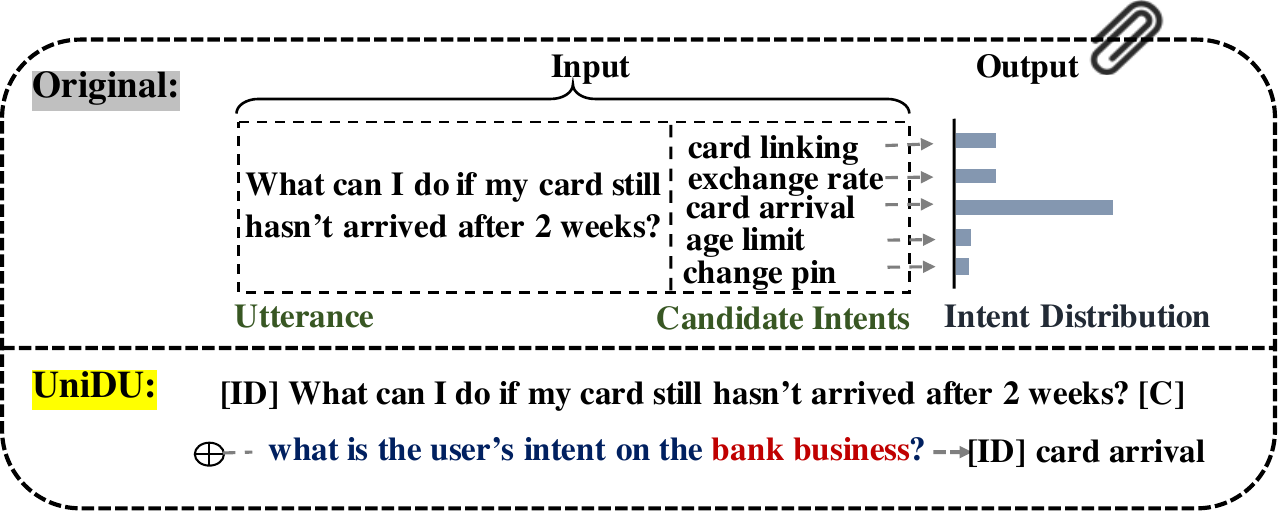} 
\end{figure}
\noindent where we do not list all the intents. To integrate the generalization capability into the UniDU model, we also construct negative samples for the intent detection task. The intent name of the negative sample is ``not defined'', where the input utterances $U_n$ are sampled from out-of-domain dialogues. The ratio of negative and positive samples is set to 2:1. 
Until now, all the five dialogue understanding tasks have been formulated as the unified sequence-to-sequence generation task. The specific examples are shown in Figure~\ref{unidu}.

\section{Multitask Training Strategies}
\label{sec:mtl}
Although the five DU tasks can be formulated as a unified generative task, straightforward multitask training may not work due to the different natures of these tasks. In this section, we discuss multitask training strategies and propose a novel model-agnostic adaptive weighting strategy. 

\subsection{Multitask Learning Classification}
The existing multitask training strategies can be classified into three categories: average sum method, manual scheduled method, and learnable weight method.

\noindent \textbf{Average Sum} method distributes all the samples with the same weight. In other words, the losses from different samples are directly averaged, formulated as $\mathcal{L} = \frac{1}{T}\sum_{t=1}^{T}\mathcal{L}_t$, 
where $T$ is the number of the tasks and $\mathcal{L}_t$ is the loss of the $t$-th task.

\noindent \textbf{Manual Schedule} method designs a heuristic training schedule for planning the learning process of different tasks. For example, curriculum learning~\cite{bengio2009curriculum} is a kind of typical manual scheduled method, which first trains the easier samples and then adds the more complicated cases. The manual scheduled method can be formulated as $\mathcal{L} = \frac{1}{\sum\mathbb{I}(t)}\sum_{t=1}^{T}\mathbb{I}(t)\cdot \mathcal{L}_t$,
where $\mathbb{I}(t)$ is indicator function, whose value is 0 or 1.

\noindent \textbf{Learnable Weight} method aims to parameterize the loss weights of different tasks. The target of the parameterized weights is to balance the effects of task instances, which prevents the model from slanting to one or several tasks and achieves global optimization. There are two classical learnable weight algorithms: homoscedastic uncertainty weighting (\textbf{HUW})~\cite{kendall2018multi} and gradient normalization (\textbf{GradNorm})~\cite{chen2018gradnorm}. For the tasks, the loss function is formulated as $\mathcal{L} = \sum_{t=1}^{T}W_t\cdot \mathcal{L}_t$,
where $W_t$ is learnable weights and greater than 0. In the HUW algorithm, the weights update as the following loss function:
\begin{equation}
    \mathcal{L}_{\rm HUW} = \sum_{t=1}^{T}(\mathcal{L}_t\cdot W_t - \log(W_t)),
\label{eq:huw}
\end{equation}
where $\log(W_t)$ is to regularize weights, which is adaptive to regression tasks and classification tasks. The motivation of the GradNorm method is to slow down the learning scale of the task that has a larger gradient magnitude and faster convergence rate. 


\subsection{Model-Agnostic Training Strategy}
\label{sec:MATS}
In Equation~\ref{eq:huw}, the learnable weight $W_t$ is only dependent on the corresponding task. Thus, we can regard the weight as the function of task $W_{\phi}(t)$, where $\phi$ are parameters shared among five tasks. Under the UniDU framework, five tasks share the same encoder-decoder model, which is a constant in weight function $W_{\phi}(t)$. The task format depends on task attributes, such as input, output, and data scale. To extract the characters of five tasks, we manually design a vector as the task feature to represent a task. Each dimension in the task feature has its physical meaning related to the model-agnostic setting. In this paper, we design 14 dimensional vector $\mathbf{f}_t$ for each task introduced in detail in Appendix~\ref{sec:apdxb}. Since the model-agnostic training strategy (MATS) formulates the weight as the task-related function and may share the function parameters among different tasks, the weights are no longer independent as in the original learnable weight method. The MATS improved from Equation~\ref{eq:huw} is formalized as:
\begin{equation}
    \mathcal{L}_{\rm MATS} = \sum_{t=1}^{T}(\mathcal{L}_t\cdot W_{\phi}(\mathbf{f}_t) - \log(W_{\phi}(\mathbf{f}_t))).
\label{eq:mats}
\end{equation}

\section{Experiments}
We conduct the experiments on ten dialogue understanding corpora. Each task has two corpora. We evaluate the UniDU framework with eight different training strategies. Compared with well-designed models, our proposed UniDU can get better performance in five benchmarks. Then we deeply analyze different factors affecting the UniDU model's performance, including DU tasks, unified format, and pre-trained language models. Last but not least, we conduct few-shot experiments to validate the generalization ability of UniDU.

\begin{table*}[h!]
\setlength\tabcolsep{5pt}
\centering
\newcolumntype{s}{>{\columncolor[HTML]{DCDCDC}} c}
\begin{tabular}{c c c c c c c c c c c c} 
 \bottomrule
  \hline
 \multirow{2}{3em}{\centering \textbf{Methods}} & \multicolumn{2}{c}{DS{\tiny (SAMSUM)}} & & \multicolumn{2}{c}{DC{\tiny (TASK)}} & & \multicolumn{1}{c}{ID{\tiny (BANKING77)}} & & \multicolumn{1}{c}{SF{\tiny (RESTAURANTS8K)}} & & \multicolumn{1}{c}{DST{\tiny (WOZ2.0)}}  \\ 
 \cline{2-3} 
 \cline{5-6}
 \cline{8-8}
 \cline{10-10}
 \cline{12-12}
 & R-1 & R-L & & EM & BLEU & & ACC. & & $F_1$ & & JGA \\
 \hline
\multirow{2}{3em}{\centering Baselines} & 49.67$^{*}$ & 48.95$^{*}$ & & 74.2 & 89.4 & & 93.44 & & 96.00 & & 91.4 \\
 & \multicolumn{2}{c}{\tiny ~\cite{wu2021controllable}} & & \multicolumn{2}{c}{\tiny ~\cite{Chen2021DecoupledDM}} & & \multicolumn{1}{c}{\tiny ~\cite{mehri2020dialoglue}} & & \multicolumn{1}{c}{\tiny ~\cite{coope2020span}} & & \multicolumn{1}{c}{\tiny ~\cite{tian2021amendable}}  \\ 
 \hline
 \multicolumn{12}{c}{Eight Training Strategies under UniDU Framework} \\
 \hline
 \textbf{ST} & 49.74 & 47.10 & & 76.4 & 89.0 & & 91.49 & & 95.76 & & 89.8 \\
 \textbf{TT} & 51.24 & 48.59 & & 76.1 & 89.2 & & 91.94 & & 95.12 & & 91.0 \\
 \hline
 \textbf{MIX} & 50.98 & 48.13 & & 76.2 & \underline{90.8} & & 91.91 & & 96.43 & & 90.8 \\
 \textbf{G2S} & 51.13 & \underline{48.75} & & 76.3 & 90.1 & & 90.12 & & 94.81 & & 86.8 \\
 \textbf{CL} & 51.04 & 48.36 & & 77.2 & 89.8 & & 92.17 & & 96.02 & & 90.8 \\ \hline
 \textbf{GradNorm} & 51.33 & 48.69 & & 77.4 & 90.4 & & 92.07 & & 96.69 & & 90.5 \\
 \textbf{HUW} & 50.31 & 47.69 & & 76.2 & 90.4 & & 93.14 & & 97.43& & 91.9 \\
 \textbf{MATS} & 50.53 & 47.97 & & 76.6 & 90.6 & & \textbf{\underline{93.60}} & & \textbf{\underline{97.61}} & & \textbf{\underline{92.3}} \\
 \hline
 \quad \textbf{Finetune} & 51.93 & \textbf{49.01} & & 76.1 & \textbf{91.0} & & 93.54 & & 97.19 & & 92.1 \\
 \hline
\end{tabular}
\caption{The results on five DU tasks trained with eight learning strategies. \textbf{Finetune} means that the best model (according to underlined metric values) of each task continues to be fine-tuned on separate task corpus. $^{*}$ means that we run their released code with BART-base instead of BART-large to fairly compare with our model.}
\label{tab:mtl}
\vspace{-5mm}
\end{table*}

\subsection{Corpora\&Metrics}
There are ten dialogue understanding corpora in total spanned five tasks: dialogue summary (DS), dialogue completion (DC), slot filling (SF), intent detection (ID), and dialogue state tracking (DST). We choose two well-studied corpora for each task: one is the evaluation corpus, and the other is the auxiliary corpus. The dataset statistics are shown in Appendix~\ref{sec:apdxa}.

\noindent \textbf{Dialogue Summary}: We choose SAMS{\small UM}~\cite{gliwa2019samsum} and D{\small IALOG}S{\small UM}~\cite{chen2021dialogsum} datasets. The common metrics for the summary task are ROUGE scores, which measure the overlap of $n$-grams in the generated summary against the reference summary.

\noindent \textbf{Dialogue Completion}: T{\small ASK}~\cite{quan2019gecor} and C{\small ANARD}~\cite{elgohary2019can} are used. The metrics are BLEU score and exact match (EM) accuracy. BLEU measures how similar the rewritten sentences are to golden ones. Exact match means the rate of the generated totally equaled to the golden.

\noindent \textbf{Intent Detection}: We conduct the experiments on B{\small ANKING77}~\cite{casanueva2020efficient} and H{\small WU64}~\cite{liu2019benchmarking}, where 77 and 64 means the number of predefined intents. The evaluation metric is detection accuracy (ACC.).

\noindent \textbf{Slot Filling}: We choose to conduct the experiments on R{\small ESTAURANTS8K}~\cite{coope2020span} and S{\small NIPS}~\cite{coucke2018snips}. We report $F_1$ scores for extracting the correct span per user utterance. Note that the correct predictions on negative samples are not calculated in the $F_1$ score, which is comparable with traditional methods.

\noindent \textbf{Dialogue State Tracking}: W{\small OZ2.0}~\cite{wen2017network} and M{\small ULTI}W{\small OZ2.2}~\cite{zang2020multiwoz} are used. The metric is joint goal accuracy (JGA), which measures the percentage of success in all dialogue turns, where a turn is considered a success if and only if all the slot values are correctly predicted. Note that we only use ``hotel'' domain data of M{\small ULTI}W{\small OZ2.2} in the training phase.

\subsection{Eight Training Strategies}
As introduced in Section~\ref{sec:mtl}, the multitask training strategies can be divided into three categories: average sum, manual schedule, and learnable weight. Before introducing MTL training methods, there is an intuitive baseline trained on its own data named single training (\textbf{ST}). In ST, the sequence-to-sequence models are only trained on five evaluated datasets, respectively.
In average sum method, there are two types of training strategies: task transfer learning (\textbf{TT})~\cite{torrey2010transfer,ruder2019transfer} and mixture learning (\textbf{MIX})~\cite{wei2021finetuned}. The task transfer learning aims to enhance the performance using external data from the auxiliary corpus that has the same task setup. This is the main reason that we select two corpora for each task. The mixture learning directly mixes up all the training samples from ten corpora together. In these two methods, the learning weight for each sample is equally distributed.
In the manual schedule method, we test two training routes according to the curriculum learning method. From the input perspective, five tasks can be divided into three classes: utterance-level input on intent detection and slot filling, turn-level input on dialogue completion, and dialogue state tracking and dialogue-level input on dialogue summary. The inputs gradually become more complex in order: utterance-level, turn-level, and dialogue-level. Thus, the intuitive method (named \textbf{CL}) trains five tasks in this order. Note that the previous data are kept in the next training phase. From the task setup perspective, dialogue summary and dialogue completion belong to domain-independent tasks. The other three tasks are domain-dependent tasks. There is another training route (\textbf{G2S}): from general tasks to domain-specific tasks.
In learnable weight method, we evaluate three methods introduced in Section~\ref{sec:mtl}: \textbf{GradNorm}, \textbf{HUW} and our proposed \textbf{MATS}. 

\subsection{Experimental Setup}
In this paper, we set BART-base as the backbone of the unified encoder-decoder model. The BART model is implemented with HuggingFace library~\cite{wolf2019huggingface}. We conduct all the experiments on the 2080TI GPU with 11G memory. we run every experiment for 60 epochs spent 72 hours. The batch size is 32 with the gradient accumulation strategy (updated per 8 steps). The learning rates of the unified model and learnable weights are 1e-5 and 1e-4, respectively. In the MATS method, the weight function consists of two linear layers with the ReLU activation function, whose hidden sizes are 64.

\subsection{Results}
In Table~\ref{tab:mtl}, we report the best evaluation performance on five tasks with eight training strategies. The well-designed models as baselines are introduced in Section~\ref{sec:intro}. The experimental results show that different training strategies greatly affect the performance of five tasks under the UniDU framework. Our proposed MATS achieves the best or near best performance except on dialogue summary. On the atypical generation tasks (intent detection, slot filling, and dialogue state tracking), the UniDU with MATS methods can achieve promising improvement compared to well-designed models. The simple task transfer learning method (TT) can not largely increase the performance compared with single training. 
The mixture operation leads to consistent performance improvement on five tasks. 
However, compared with TT, the improvement is still limited except for dialogue completion. 
Compared with our proposed MATS, MIX biases convergence to more complex DU tasks (dialogue summary and dialogue completion). 
Two manual schedule methods (G2S and CL) do not have any distinct advantages. In learnable weight methods, GradNorm only achieves excellent performance on dialogue summary. HUW achieves performance gain on intent detection, slot filling, and dialogue state tracking. We continue fine-tuning the best UniDU models (signed with underline) on the corresponding corpus. We find that only the dialogue summary and dialogue completion have obvious performance gain, which reflects the necessity of the UniDU framework for simpler generative tasks.

\begin{table}[t]
\setlength\tabcolsep{3pt}
\centering
\newcolumntype{s}{>{\columncolor[HTML]{DCDCDC}} c}
\begin{tabular}{ccccccc} 
 \bottomrule
  \hline
 \multirow{2}{3em}{\centering \textbf{Methods}} & \multicolumn{1}{c}{DS} & \multicolumn{1}{c}{DC} & \multicolumn{1}{c}{ID} & \multicolumn{1}{c}{SF} & \multicolumn{1}{c}{DST} & \multirow{2}{3em}{\centering Overall} \\ 
 & {\tiny (R-L)} & {\tiny (BLEU)} & {\tiny (ACC.)} & {\tiny ($F_1$)} & {\tiny (JGA)} & \\
 \hline
 \textbf{MIX} & \textbf{48.04} & 90.40 & 91.9 & 96.43 & 90.1 & 83.23 \\
 \textbf{HUW} & 47.63 & 89.95 & 93.0 & 97.43 & 91.8 & 83.97 \\
 \textbf{MATS} & 47.57 & \textbf{90.43} & \textbf{93.5} & \textbf{97.46} & \textbf{91.9} & \textbf{84.16} \\
 \hline
\end{tabular}
\caption{The best overall performance of MIX, HUW and MATS methods.}
\label{tab:ft}
\vspace{-3mm}
\end{table}

\begin{table}[t]
\setlength\tabcolsep{0.3pt}
\centering
\newcolumntype{s}{>{\columncolor[HTML]{DCDCDC}} c}
\begin{tabular}{lccccc} 
 \bottomrule
 \multirow{2}{3em}{\centering \textbf{Method}} & DS & DC & ID & SF & DST \\ 
 & {\tiny (R-L)} & {\tiny (BLEU)} & {\tiny (ACC.)} & {\tiny ($F_1$)} & {\tiny (JGA)} \\
 \hline
\textbf{MATS} & 47.97 & 90.6 & 93.60 & 97.61 & 92.3 \\
 \hline
  \textbf{- DS} & - & 90.2{\tiny \color{red}{$\blacktriangledown$}0.4} & 93.20{\tiny \color{red}{$\blacktriangledown$}0.4} & 97.35{\tiny \color{red}{$\blacktriangledown$}0.26}  & 92.8{\tiny \color{green}{$\blacktriangle$}0.5} \\
  \textbf{- DC} & 47.77{\tiny \color{red}{$\blacktriangledown$}0.20} & - & 93.41{\tiny \color{red}{$\blacktriangledown$}0.19} & 97.39{\tiny \color{red}{$\blacktriangledown$}0.22} & 91.8{\tiny \color{red}{$\blacktriangledown$}0.5} \\
  \textbf{- ID} & 47.81{\tiny \color{red}{$\blacktriangledown$}0.16} & 90.5{\tiny \color{red}{$\blacktriangledown$}0.1} & - & 97.45{\tiny \color{red}{$\blacktriangledown$}0.16} & 92.3{\tiny <0.0} \\
  \textbf{- SF} & 47.77{\tiny \color{red}{$\blacktriangledown$}0.20} & 90.5{\tiny \color{red}{$\blacktriangledown$}0.1} & 93.60{\tiny <0.0} & - & 92.0{\tiny \color{red}{$\blacktriangledown$}0.3} \\
  \textbf{- DST} & 47.85{\tiny \color{red}{$\blacktriangledown$}0.12} & 90.6{\tiny <0.0} & 93.47{\tiny \color{red}{$\blacktriangledown$}0.13} & 97.58{\tiny \color{red}{$\blacktriangledown$}0.03} & - \\
 \hline
\end{tabular}
\caption{Ablation study on effects of each task corpora.}
\label{tab:no_summary}
\vspace{-5mm}
\end{table}

In Table~\ref{tab:mtl}, we report the task-specific performance of the UniDU model, whose checkpoints are selected by the task-specific metric. Table~\ref{tab:ft} shows unified performance on five tasks with MIX, HUW, and MATS methods. We evaluate the single checkpoint of UniDU model, which has the highest evaluated overall score, on the five tasks. The overall score is the average value of the five main metrics shown in Table~\ref{tab:ft}. We can see that our proposed MATS gets the highest overall performance and the best performance on four DU tasks.

\subsection{Analysis}
In this subsection, we analyze factors to affect the performance of UniDU model including DU tasks, unified format and pre-trained language models.
\begin{table}[t]
\setlength\tabcolsep{4pt}
\centering
\newcolumntype{s}{>{\columncolor[HTML]{DCDCDC}} c}
\begin{tabular}{cccccc} 
 \bottomrule
 \multirow{2}{3em}{\centering \textbf{Backbone}} & DS & DC & ID & SF & DST \\ 
 & {\tiny (R-L)} & {\tiny (BLEU)} & {\tiny (ACC.)} & {\tiny ($F_1$)} & {\tiny (JGA)} \\
 \hline
\textbf{Trans.-B} & 34.84 & 74.2 & 86.36 & 83.01 & 72.5 \\
\textbf{BART-B} & \textbf{47.97} & \textbf{90.6} & \textbf{93.60} & \textbf{97.61} & \textbf{92.3} \\
\textbf{T5-S} & 41.63 & 85.9 & 87.04 & 96.94 & 89.9 \\
 \hline
\textbf{Trans.-L} & 34.10 & 67.4 & 86.46 & 71.65 & 71.0 \\
\textbf{BART-L} & \textbf{48.89} & 88.6 & 93.44 & 97.12 & \textbf{92.6} \\
\textbf{T5-B} & \textbf{48.89} & \textbf{90.7} & \textbf{93.90} & \textbf{98.14} & \textbf{92.6} \\
 \hline
\end{tabular}
\caption{Ablation study on effects of different pre-trained language models with encoder-decoder architecture. 
}
\label{tab:ablation_plm}
\vspace{-6mm}
\end{table}

\subsubsection{Effects of DU Tasks}
To validate the effects of the dialogue understanding tasks, we directly remove one of five DU corpora and train the UniDU model with the MATS method shown in Table~\ref{tab:no_summary}. 
In general, the five DU tasks benefit each other, except that dialogue summary has negative effects on the dialogue state tracking task. 
We guess the general dialogue summary task summarizes a dialogue into a sentence, ignoring the domain-specific information. On the other hand, we find that the dialogue completion task has the most significant effect on the other four DU tasks. 
It indicates that the co-reference and information ellipsis are still the main factors to impact the dialogue understanding ability. 
The phenomenon can facilitate the dialogue understanding community to pay more attention to dialogue completion. For example, when pre-training a scaling dialogue model, the pre-trained tasks should be close to the dialogue completion task. 

\begin{table*}[t]
\setlength\tabcolsep{2pt}
\centering
\small
\begin{tabular}{cc}
  \hline
  \textbf{Unseen Dialogue Content} & \textbf{UniDU{\tiny MATS}} \\
    \hline
    \makecell[l]{[DS] USER : I'd like a taxi to \textbf{\color{blue}{take me to ruskin gallery}} [T] SYSTEM : Sure! What is your \\ departure site? [T] USER : I will depart from saffron brasserie \textbf{\color{blue}{at 7:15}}. What is the car type \\ and contact number so I know who and where you will pick me up? [T] SYSTEM : Booking \\ completed! \textbf{\color{blue}{A grey ford}} will be picking you up. The contact number is 07689877132. [T] \\ USER : That is all I needed, thank you. [C] what’s the summary of this dialogue?} & \makecell[l]{[DS] a grey ford will take \\ USER to ruskin gallery at \\ 7:15.} \\
    \hline
    \makecell[l]{[DC] USER : Please reserve for me a taxi that will \textbf{\color{blue}{pick me up at cambridge arts theatre}} after \\ 09:30 [T] SYSTEM : And where will you be going? [T] USER : I'm going to restaurant one \\ seven. [T] SYSTEM : Your booking is complete, \textbf{\color{blue}{a black audi will be picking you up}}. [T] \\ USER : Thank you. \textbf{\color{blue}{I need the contact number}}, as well. [C] what is the semantic completion \\ statement of ``Thank you. I need the contact number, as well.''?} & \makecell[l]{[DC] I need the contact \\ number of a black audi to \\ pick me up at cambridge \\ arts theatre} \\
    \hline
    \makecell[l]{[ID] help me \textbf{\color{blue}{get a taxi to the cambridge museum of technology}} please. [C] what is the user's \\ intent on the taxi?} & \makecell[l]{[ID] transport taxi} \\
    \hline
    \makecell[l]{[SF] I need a taxi to pick me up at Ashley Hotel to \textbf{\color{blue}{leave after 10:45}}. [C] what is leaving \\ time of taxi?} & \makecell[l]{[SF] 10:45} \\
    \hline
    \makecell[l]{[DST] USER : I need a taxi. I am  \textbf{\color{blue}{going to avalon}} and I need to leave after 16:15 [C] what \\ is the user's constraint about the destination of the taxi?} & \makecell[l]{[DST] avalon} \\
    \hline
  \end{tabular}
\caption{Case study of the zero-shot performance of the best unified model trained with MATS method. The input dialogue contents are sampled from unseen ``Taxi'' domain.}
\label{tab:case}
\vspace{-3mm}
\end{table*}

\subsubsection{Effects of Unified Format}
\begin{figure}[t]
\centering
\includegraphics[width=0.38\textwidth]{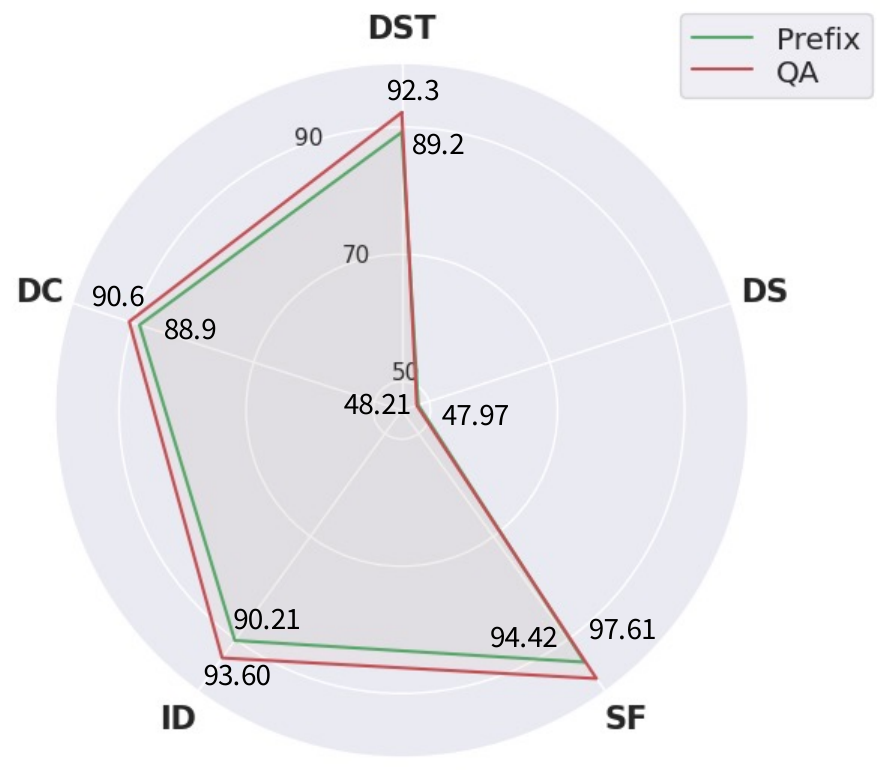} 
\caption{Ablation study of different unified understanding format.}
\label{format}
\vspace{-3mm}
\end{figure}

As introduced in Section~\ref{sec:unidu}, we formulate dialogue understanding tasks in QA format. There is an intuitive alternative: prefix format, where the task query is concatenated on the decoder side. At inference time, the decoder is directly fed with task query and then generates the answer. As shown in Figure~\ref{format}, the QA format achieves a performance boost on four of five DU tasks (except for dialogue summary) compared to the prefix format.

\subsubsection{Effects of PLMs}
To validate the effects of the different pre-trained backbones, we initialize the encoder-decoder of UniDU model with random mechanism, BART~\cite{lewis2020bart} and T5~\cite{raffel2020exploring}. The \textbf{Trans.-B} and \textbf{Trans.-L} in Table~\ref{tab:ablation_plm} mean the random-initialized Transformer trained from scratch, which has the same parameters with BART-base model (\textbf{BART-B}) and BART-large model (\textbf{BART-L}). \textbf{T5-S} and \textbf{T5-B} mean T5-small and T5-base respectively. We can see that the pre-trained language models get absolute performance gain compared to random-initialized models. BART-B can get better performance than T5-S. When the parameter scale increases, T5-base achieves the best performance than other models. The results show that the large PLMs can improve the complex dialogue summary by a large margin.

\subsection{Generalization Ability}
To further evaluate the generalization ability of the UniDU model, we first conduct few-shot learning experiments on the domain-dependent slot filling task. We test the zero-shot capability of UniDU on unseen dialogue data.

\noindent \textbf{Few-shot Learning:} 
We select the UniDU model that gets the best evaluation of overall performance on five tasks learned with the MATS method. For the slot filling task, we extend another dialogue corpus D{\small STC8}~\cite{rastogi2020schema}. We choose the ``Bus'' domain data in D{\small STC8}, which is unseen in the training process of UniDU. 
Compared with vanilla BART, UniDU has obvious advantages, especially in the extremely resource-limited situation. When there is only 1\% training data, the vanilla BART is disabled to learn, as shown in Figure~\ref{few}. The few-shot experiment on the DST task is shown in Appendix~\ref{sec:apdxc}.


\noindent \textbf{Zero-shot Performance:} We validate UniDU model trained with MATS method on unseen ``Taxi'' domain dialogue data collected from M{\small ULTI}W{\small OZ2.2} corpus. UniDU model can get 18.24\% accuracy on ID, 39.69\% F1 score on SF and 1.6\% JGA on DST.



\begin{figure}[t]
\centering
\includegraphics[width=0.38\textwidth]{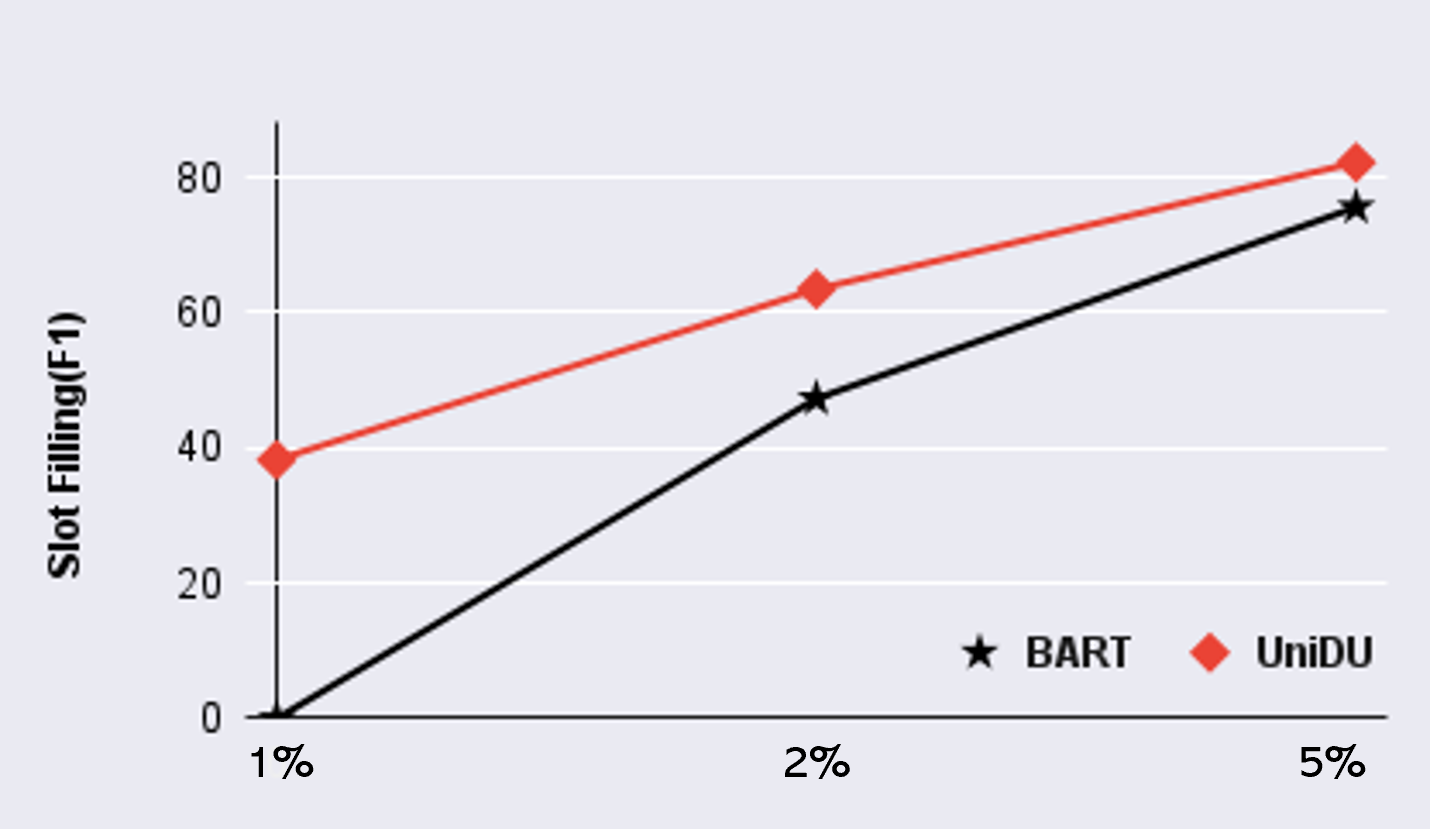} 
\caption{Few-shot learning results on slot filling fine-tuned on BART and UniDU. 1\%, 2\% and 5\% are the percents of the training data on unseen ``Bus'' domain.}
\label{few}
\vspace{-3mm}
\end{figure}

\section{Case Study}
\label{sec:apdxd}
 We directly validate the UniDU model trained with the MATS method on unseen ``Taxi'' domain dialogue data collected from M{\small ULTI}W{\small OZ2.2} corpus. As shown in Table~\ref{tab:case}, we find that the UniDU model can generate reasonable dialogue summary and completion. Note that the UniDU model did not see any task-oriented dialogue in these two tasks. For domain-specific tasks, the UniDU model can still generate accurate query answers in some cases. It indicates that our proposed generative UniDU model has excellent generalization ability, which not only can adapt to unseen dialogue and also directly generate reasonable answers on five DU tasks in the zero-shot setting.
 
 \begin{figure}[t]
\centering
\includegraphics[width=0.38\textwidth]{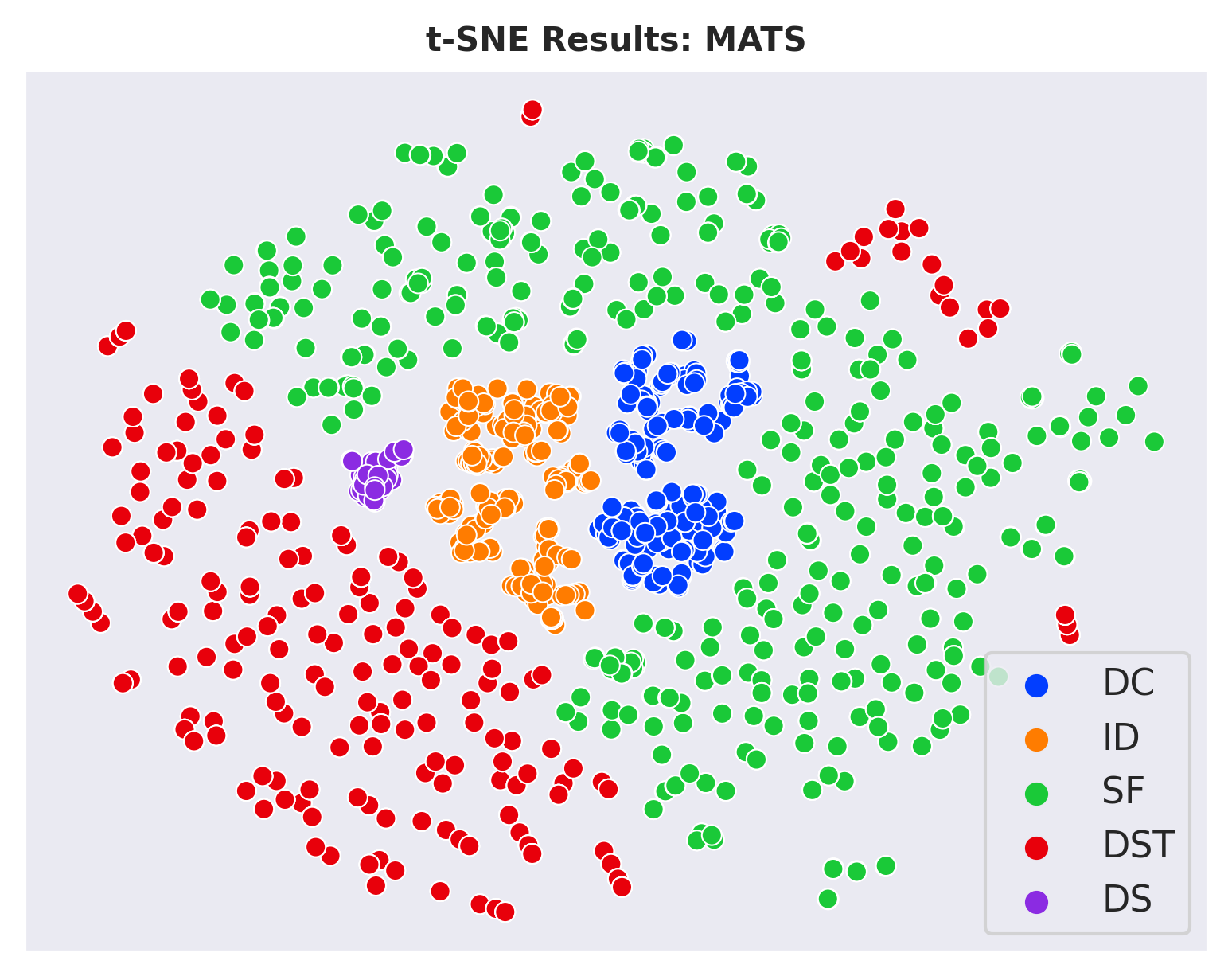} 
\caption{The reduce-dimension map of task embeddings collected from UniDU model trained by MDTS. The task embedding is the final decoder representation of the task identification token.}
\label{fig:reduce}
\vspace{-4mm}
\end{figure}
 
 To further explore the relations among five tasks, we plot the reduced-dimension map of the task embeddings of five tasks with the t-SNE algorithm shown in Figure~\ref{fig:reduce}. The task embeddings are the final decoder layer representation of the task identification token, whose model is trained with MDTS. The dialogue data is from the above unseen ``Taxi'' domain to eliminate the impacts of the dialogue context. We find that the embeddings of dialogue summary, dialogue completion, and intent detection cluster together. These three tasks under the UniDU framework are more general than slot filling and dialogue state tracking, whose task queries are slot-wise. The task formats between slot filling and dialogue state tracking are close. However, the UniDU model can still have good performance to distinguish between these two tasks.

\section{Related Work}
Our work relates to several broad research areas, including prompting, dialogue modeling, and multitask learning. Due to the content limitation, here we describe one subarea: multitask learning in NLP applications that relate most closely to our work. ~\citet{luong2016mtl} apply a sequence-to-sequence model on three general NLP tasks and study different parameter-sharing strategies. ~\citet{kumar2016ask,mccann2018natural} try to cast NLP tasks as QA over a context. The main topics in this work are how to design an efficient model to integrate the knowledge between question and context. \citet{liu2019multi} combine four natural language understanding tasks, which utilize BERT as the shared representation model. The model corresponding to each task still has a well-designed part of solving the intrinsic problem. It hampers the analysis of the interaction among the different tasks.

Recently, \citet{wei2021finetuned} formulated the NLP tasks as the generation task by directly mixing scaling annotated data up. They only focus on zero-shot and few-shot ability on the NLP tasks and ignore the impacts of the different multitask training strategies, which can not achieve better performance on general NLP tasks compared to supervised learning methods on well-designed models. In task-oriented dialogue (TOD) modelling, ~\citet{peng2020soloist,su2021multi} reformulate the pipeline TOD model as the sequential end-to-end generation problem. The end-to-end model needs to generate dialogue state, dialogue action, and response at the same time, which is not scalable when the number of tasks increases. The sequential format needs all the annotations of the same context, which is unavailable in the DU area. Most recently, PPTOD~\cite{su2021multi} unifies the TOD task as multiple generation tasks, including intent detection, DST, and response generation. However, they focus on the response generation ability and ignore the effects of different tasks. In this paper, we deep dive into analyzing the effects of five DU tasks. 

\section{Conclusion\&Future Work}
In this paper, we propose a unified generative dialogue understanding framework (UniDU) to share the knowledge across five typical dialogue understanding tasks. We introduce a model-agnostic adaptive weight learning method for multitask training to alleviate the biased generation problem. Our proposed UniDU method achieves better performance compared to well-designed models on a total of five DU tasks. We further deep dive into studying the affected factors. Finally, experimental results indicate that our proposed UniDU model can also get excellent performance under few-shot and zero-shot settings. In the future, we will increase the scale of the DU corpora and integrate the unsupervised dialogue pre-training tasks. We will further examine the task-level transferability of the UniDU model.

\section*{Acknowledgements}
We sincerely thank the anonymous reviewers for their valuable comments. We thank the SIGDIAL mentors Stefan Ultes and Ondrej Dusek to help us prepare our final submission. This work has been supported by the China NSFC Projects (No.62120106006 and No. 62106142), Shanghai Municipal Science and Technology Major Project (2021SHZDZX0102), CCF-Tencent Open Fund and Startup Fund for Youngman Research at SJTU (SFYR at SJTU).


\bibliography{anthology,custom}
\bibliographystyle{acl_natbib}

\clearpage
\appendix

\begin{center}
    \Huge \textbf{\textup{Appendix}}
\end{center}
\vspace{2mm}
\hrule height 0.07cm 

\section{Dialogue Understanding Corpora}
\label{sec:apdxa}

\begin{table}[h!]
\setlength\tabcolsep{2pt}
\centering
\newcolumntype{s}{>{\columncolor[HTML]{DCDCDC}} c}
\begin{tabular}{cccccc} 
 \bottomrule
  \hline
 \textbf{Corpora} & \textbf{\#Sample} & \textbf{I{\tiny (Token)}} & \textbf{I{\tiny (Turn)}} & \textbf{O{\tiny (Token)}} & \textbf{Task}\\
 \hline
 SAMS{\tiny UM} & 14732 & 104.95 & 11.16 & 20.31 & DS  \\
 D{\tiny IALOG}S{\tiny UM} & 12460 & 140.48 & 9.49 & 22.86 & DS  \\
 \hline
 T{\tiny ASK} & 2205 & 34.92 & 2.75 & 10.84 & DC  \\
 C{\tiny ANARD} & 31526 & 102.67 & 9.80 & 11.55 & DC  \\
 \hline
 B{\tiny ANKING77} & 12081 & 21.64 & 1 & 3.14 & ID  \\
 H{\tiny WU64} & 25715 & 17.69 & 1 & 2.05 & ID  \\
 \hline
 R{\tiny ESTAURANTS8K} & 15270 & 14.44 & 1 & 3.38 & SF  \\
 S{\tiny NIPS} & 35748 & 15.31 & 1 & 1.77 & SF  \\
 \hline
 W{\tiny OZ2.0} & 7608 & 78.96 & 4.63 & 1.30 & DST  \\
 M{\tiny ULTI}W{\tiny OZ2.2} & 35119 & 115.80 & 5.99 & 1.45 & DST  \\
 \hline
\end{tabular}
\caption{The ten DU corpora trained on UniDU model. \textbf{I{\tiny (Token)}} and \textbf{I{\tiny (Turn)}} mean the average length of the split tokens and the average turns of the input dialogue content. \textbf{O{\tiny (Token)}} means the average length of the split tokens of the task-specific output.}
\label{tab:corpora}
\end{table}

\noindent In this paper, we train our proposed unified generative model on ten dialogue understanding corpora, as shown in Table~\ref{tab:corpora}. For each DU tasks, we select two well-studied datasets. The first one is used to evaluate and the second one is an auxiliary corpus. The main reason to select two datasets for each task is to compare the multitask learning with the task transfer learning. We aim to know whether the knowledge sharing between different dialogue understanding data is only happening in the same DU task rather than all the DU tasks. The experimental results show that the annotated data from the other DU tasks are also important to enhance the performance, which indicates that it is an efficient way to transfer the knowledge among all the DU tasks. Note that the selected DU data are from different corpora, which means that the distribution of the input dialogue content is totally different. As shown in Table~\ref{tab:corpora}, the inputs and the outputs of the five DU tasks are greatly different from each other. The longest average input reaches to 140.48 and the shortest is only 14.44. The longest output is 22.86 from dialogue summary and the shortest is 1.30 from dialogue state tracking. These characters lead a big challenge to train all the dialogue understanding data in multitask learning way. The experimental results show that the intuitive mixture learning method makes UniDU model bias convergence to the more complex tasks like dialogue summary and dialogue completion. In this paper, we compare eight multitask training strategies. Our proposed MATS method can achieve the best overall performance on the five tasks under UniDU framework.

\begin{figure}[H]
\centering
\includegraphics[width=0.45\textwidth]{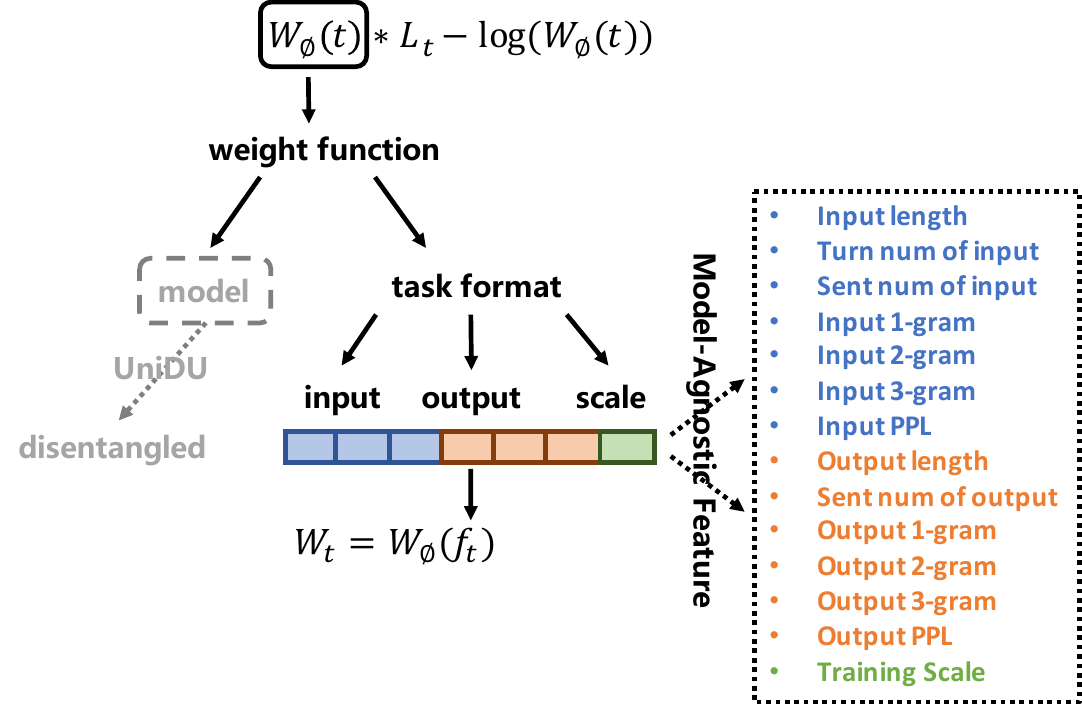} 
\caption{Overview of model-agnostic training strategy.}
\label{dm}
\end{figure}

\section{Model-Agnostic Training Strategy}
\label{sec:apdxb}
In traditional HWU algorithm, the learnable weight $W_t$ is only dependent on the corresponding task. Thus, we can regard the weight function of task $W_{\phi}(t)$, where $\phi$ are parameters shared among five tasks. Generally, the task is associated with two factors: its corresponding model and task format. Under UniDU framework, five tasks share the same encoder-decoder model, which can be regarded as a constant in weight function $W_{\phi}(t)$. The task format dependents on model-agnostic task setting, such as input, output and data scale. To distinguish the five tasks under UniDU framework, we manually design a vector as the task feature to represent a task. Each dimension in the task feature has its physical meaning related to model-agnostic setting. In this paper, we design 14 dimensional vector $\mathbf{f}_t$, as shown in Figure~\ref{dm}. For input and output, we add the average length of token, the average sentence number, the n-grams and the perplexity (PPL) as the attributes of the DU tasks. Especially for input, the average turn number is also an important character. The last attribute is training scale for each task.
Since the model-agnostic training strategy (MATS) formulates the weight as the task-related function and may share the function parameters among different tasks, the weights are not longer independent to each other as in original learnable weight method.

\begin{figure}[t]
\centering
\vspace{-5mm}
\includegraphics[width=0.4\textwidth]{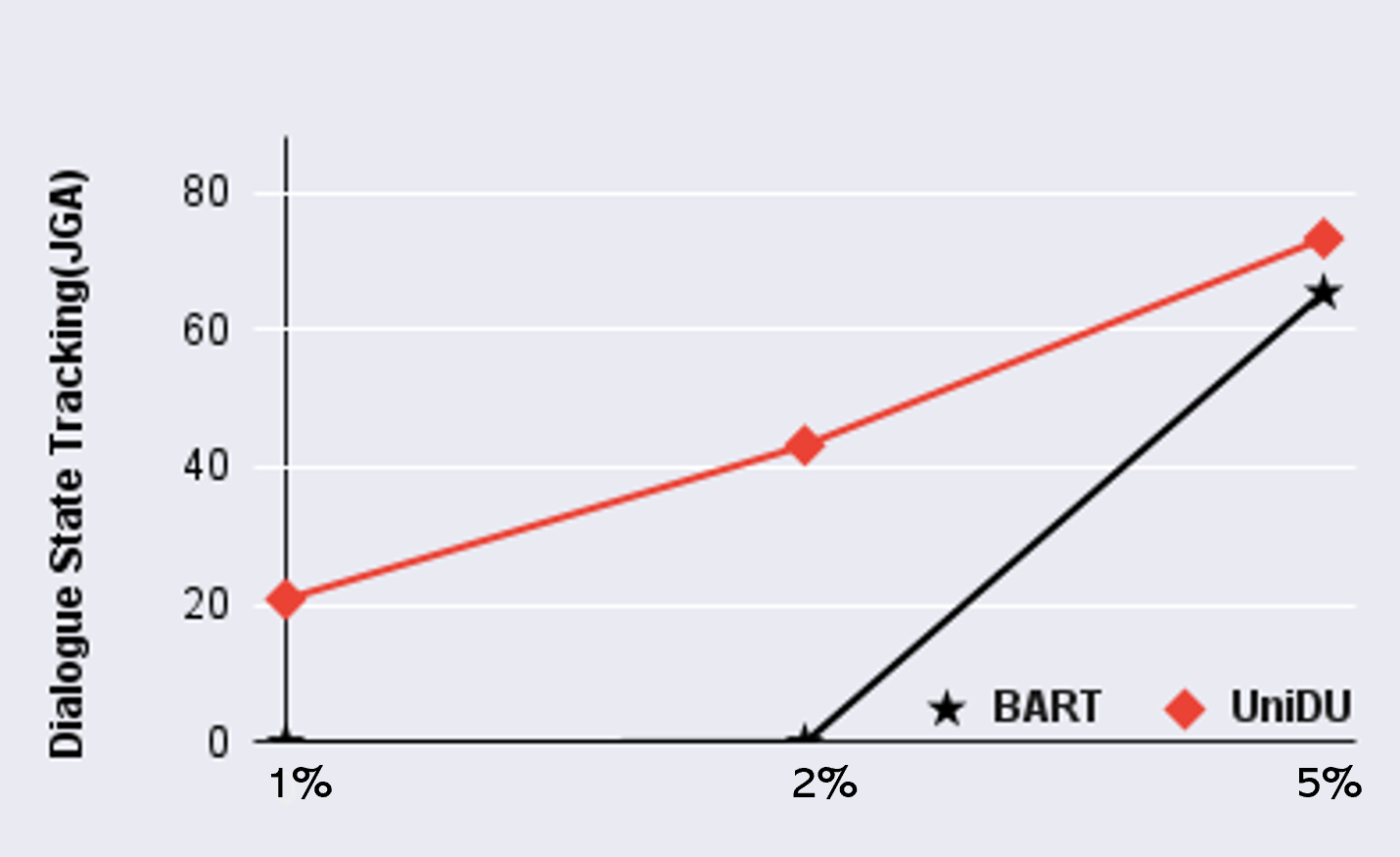} 
\caption{Few-shot learning results on DST fine-tuned on BART and UniDU. 1\%, 2\% and 5\% are the percents of the training data on unseen ``Taxi'' domain.}
\label{few_dst}
\vspace{-5mm}
\end{figure}

\section{Few-shot Learning}
\label{sec:apdxc}
We select UniDU model that gets the best evaluation overall performance on five tasks learned with MATS method.
For dialogue state tracking, we utilize the ``Train'' domain data in M{\small ULTI}W{\small OZ2.2}, which is unseen in MTL training phase.
Compared with vanilla BART, UniDU has obvious advantages, especially on extremely resource-limited situation. When there is only 1\% and 2\% training data, the vanilla BART is disable to learn. UniDU model warmed up by MATS method can quickly adapt the model on the unseen domain.

\end{document}